\documentclass{article}
\usepackage{iclr2027_conference,times}
\usepackage[T1]{fontenc}
\usepackage{amsmath,amssymb}
\usepackage{booktabs}
\usepackage{graphicx}
\usepackage{microtype}
\usepackage{tikz}
\usepackage{url}
\AtBeginDocument{\hfuzz=0pt\vfuzz=0pt}

\definecolor{capReasoning}{HTML}{FBE2E5}
\definecolor{capVisual}{HTML}{E0EFFF}
\definecolor{capText}{HTML}{FFF0CC}
\definecolor{capAction}{HTML}{E3F1E3}
\definecolor{capObject}{HTML}{EEE5F6}
\newlength{\captagwidth}
\definecolor{capAccelerate}{HTML}{DDF2EF}
\newcommand{\captag}[3][\captagwidth]{\tikz[baseline=(tag.base)]{\node[fill=#2,rounded corners=2pt,inner xsep=1.5pt,inner ysep=0.75pt,outer sep=0pt] (tag) {\footnotesize\makebox[#1][c]{\strut #3}};}}

\usepackage{caption}
\definecolor{resourceblue}{HTML}{4B7FBC}

\usepackage{wrapfig}
\usepackage{needspace}
\usepackage[colorlinks=true,allcolors=black]{hyperref}
\hypersetup{pdftitle={Agent as Policy for Robotic Manipulation},pdfauthor={Mengzhao Jia, Yang Lin, Xixin Zhang, Zhihan Zhang, Meng Jiang}}
\graphicspath{{imgs/}}
\title{Agent as Policy for Robotic Manipulation}
\author{%
Mengzhao Jia\textsuperscript{1,*} \quad
Yang Lin\textsuperscript{2,*} \quad
Xixin Zhang\textsuperscript{2,3,*} \quad
Zhihan Zhang\textsuperscript{1} \quad
Meng Jiang\textsuperscript{1} \\[0.5em]
\textnormal{\footnotesize
\textsuperscript{1}University of Notre Dame \quad
\textsuperscript{2}University of California San Diego \quad
\textsuperscript{3}San Diego State University} \\[0.3em]
\textnormal{\textsuperscript{1}\href{mailto:mjia2@nd.edu}{\texttt{mjia2@nd.edu}} \quad \textsuperscript{*}Core contributors} \\[0.3em]
{\hypersetup{urlcolor=resourceblue}\bfseries
\href{https://agent-as-policy-2026.github.io/\#top}{[Website]}\quad
\href{https://huggingface.co/datasets/Agent-as-Policy/agent-as-policy}{[Data]}\quad
\href{https://github.com/agent-as-policy-2026/agent-as-policy}{[Code]}}
}

\iclrfinalcopy
\begin{document}
\maketitle
\lhead{Agent as Policy}
\raggedbottom
\begin{abstract}
We demonstrate that a general-purpose agent can directly drive a physical robot throughout task execution without any task-specific or environment-specific training. We introduce \textbf{Agent as Policy} (AGP), which places task planning and execution under the agent's control. Given a task and a robot interface, the agent interprets visual evidence, writes executable programs, issues motion commands, and revises its actions in response to physical outcomes. This brings the agent's reasoning and programming capabilities into continuous interaction with the physical world. We study AGP across multiple real-world manipulation tasks spanning precision manipulation, dynamic motions, and deformable objects. These include assembly from human videos, block construction from goal images, dice flipping, targeted throwing, and bimanual towel folding. AGP achieves success rates of at least 80\% in seven of eight task configurations and significantly outperforms previous agentic robot systems. We further study efficiency through task experience accumulation and find that reusing saved procedures and programs shortens execution time across repeated trials. These findings support a path for general-purpose agents to act as robot policies, extending their autonomy to physical manipulation through runtime reasoning, programming, and interaction.

\end{abstract}
\section{Introduction}
\label{sec:introduction}

\begin{figure}[t]
    \centering
    \includegraphics[width=\linewidth]{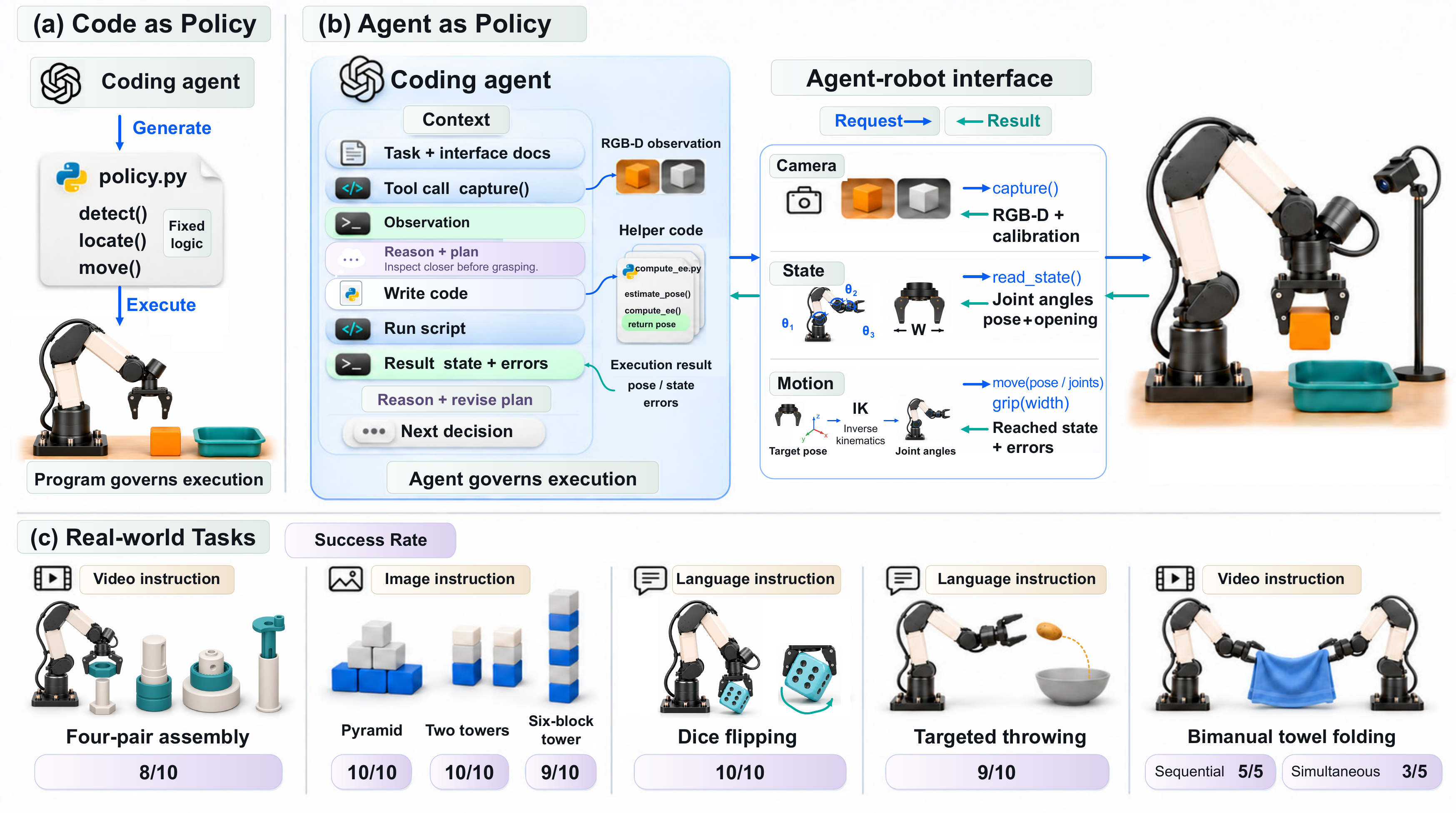}
    \caption{\textbf{Agent as Policy (AGP) for robotic manipulation.} (a) Code as Policy uses a program generated in advance to govern robot execution. (b) AGP uses the coding agent itself as the runtime policy. The agent requests observations, interprets the scene, writes and runs its own code to carry out the task, and uses execution feedback to revise its plans and subsequent actions. The robot interface provides calibrated RGB-D observations, robot state, and motion commands with execution feedback. (c) Five task groups span video, image, and language instructions, including assembly, block construction, dice flipping, targeted throwing, and bimanual towel folding. Results using GPT-6 Astra are shown as successful trials over total trials across eight task configurations.}
    \label{fig:method}
\end{figure}

Multimodal large language models (MLLMs) demonstrate strong general capabilities~\citep{li2026vlmsurvey}. Agents built on these models tackle digital tasks such as software development and computer use~\citep{yang2024sweagent,wang2024openhands,xie2024osworld}. Recent work extends these agents to robot manipulation, bringing their task execution capabilities from digital environments into the physical world~\citep{li2026roboclaw,zhang2026harnessvla,galanti2026physicalagency}. These agentic robot systems broadly follow two approaches. The first approach uses agents to generate executable programs in advance, expressed as code or computation graphs that the robot runs during task execution~\citep{liang2022code,chen2026gap}. The second approach uses agents as orchestrators of existing learned policies, selecting and sequencing these policies according to task goals and execution feedback~\citep{li2026roboclaw,zhang2026harnessvla,galanti2026physicalagency}. Both approaches, however, have limitations. Generated programs must express runtime decisions through explicit rules, making it difficult to anticipate how uncertain observations and unexpected outcomes should affect execution. Policy orchestration gives agents control over which policy runs next, while motion generation remains limited by what the selected policies can perform.

Yet physical tasks often require the flexibility to adapt both how evidence is interpreted and how actions are generated. After a failed insertion, for example, the robot may need a closer view, a revised alignment estimate, or a different approach direction. Choosing and carrying out the appropriate response requires reasoning jointly about the current scene, the attempted motion, and its outcome. To this end, we introduce \textbf{Agent as Policy (AGP)}, which uses the agent itself as the policy throughout task execution. We grant the agent full autonomy over runtime task decisions: it decides what to observe, how to interpret evidence, and which motion to request, and can author and revise local programs as needed. Each result informs the next decision, including whether to gather more evidence or change the execution strategy. This gives the agent control over perception, motion generation, and recovery at the level where new physical evidence becomes available.

% We implement AGP by connecting a general purpose agent to the robot through an interface that exposes calibrated observations, geometric queries, and motion commands (Figure~\ref{fig:method}). The coding agent provides the programming tools and persistent workspace used to create scripts, inspect data, and retain execution evidence. Through these tools, it can compute object geometry, evaluate candidate motions, and revise action sequences using feedback from the robot. The interface grounds these computations in camera calibration and robot coordinates, validates requested motion, and returns measured state and execution reports. It also supports coordinated joint and gripper motion for actions that depend on timing. Motion execution and monitoring run independently of agent inference, allowing the robot to execute a submitted motion while the agent reasons between tool calls. Adaptation occurs through updated measurements, programs, and decisions within the session, with model parameters fixed throughout execution.

We implement AGP by giving a general purpose agent an interface for observing and controlling the robot (Figure~\ref{fig:method}(b)). Guided by task instructions, interface documentation, observations, and tool results, the agent autonomously reasons about the task, plans its actions, and decides what to do next. It calls tools to request observations and writes and runs its own code to interpret visual evidence, compute motion targets, and control the robot. The interface returns camera images, robot state, and action feedback, which inform the agent's next decision. The agent decides when to gather more observations, revise its plans or code, or execute further actions. Model parameters remain unchanged throughout execution.

We evaluate AGP on real robots across five tasks: part assembly, block construction, dice flipping, targeted throwing, and bimanual towel folding. AGP achieves success rates of at least 80\% in seven of eight task configurations, including 100\% success in four. It also achieves substantially higher success rates than Graph as Policy~\citep{chen2026gap} and ASPIRE~\citep{lu2026aspire}, which use agents to build fixed robot execution pipelines rather than using an agent directly as the robot policy at runtime. We further study experience accumulation and transfer. Experience saved in previous trials improves efficiency in subsequent trials and can be transferred to a weaker model to improve its task success and efficiency. We hope this study provides insights for future research on using agents for more capable and efficient robot manipulation. Code is available here\footnote{\url{https://github.com/agent-as-policy-2026/agent-as-policy}}.

\section{Related Work}
\label{sec:related-work}

\subsection{Agentic Robot Systems}

Recent robot systems use language model agents to generate executable policies or orchestrate learned skills~\citep{chen2026gap,li2026roboclaw,liu2026roboicl}. \textbf{Program synthesis and refinement.} One line generates programs that map observations to actions through perception and control interfaces~\citep{singh2022progprompt,chen2024roboscript,mu2024robocodex}. Code as Policies~\citep{liang2022code} generates Python policy programs whose control flow remains fixed during task execution. Perception, decision making, and failure recovery follow predefined rules and conditional branches. More recently, agentic systems have extended this approach through iterative refinement using execution feedback. GaP~\citep{chen2026gap} constructs modular computation graphs and refines their structure and parameters through simulation rehearsal. ASPIRE~\citep{lu2026aspire} refines robot programs using execution feedback and reuses the refined programs for deployment. AGP differs from these program synthesis approaches by using the agent itself as the runtime policy. It interprets new images and action outcomes throughout execution and writes or revises programs to update perception, motion generation, and recovery as needed. \textbf{Runtime policy orchestration.} A second line selects and sequences learned skills according to task goals, affordances, and environment feedback~\citep{ahn2022saycan,huang2022inner}. Recent systems extend this approach to vision language action (VLA) policies with progress monitoring and recovery~\citep{li2026roboclaw,zhang2026harnessvla,galanti2026physicalagency}. These systems assign motion generation to the selected policy or primitive. AGP lets the runtime agent compute motion targets through programming and geometric reasoning, supporting task execution without a separately trained action policy.

\subsection{Robot Manipulation}

Classical robot manipulation combines geometric modeling, task and motion planning, and feedback control~\citep{garrett2021integrated}. Learned visuomotor policies predict actions from robot trajectory data, with diffusion models capturing action sequences~\citep{chi2023diffusion}. Vision language action models extend this approach to diverse tasks through image and language conditioning~\citep{brohan2023rt2,kim2024openvla,black2024pi0}. Video and world action models further couple actions with future visual states to inform action generation and evaluation~\citep{li2025uva,ye2026worldaction}. AGP introduces an alternative to policies trained specifically for action generation by using a general purpose foundation multimodal large language model (MLLM) directly as the robot policy.

\subsection{Agentic Workflows Beyond Robotics}

Language agents combine reasoning, tool use, and reflection on interaction feedback~\citep{yao2022react,schick2023toolformer,shinn2023reflexion}. Coding agents apply these mechanisms to editing and testing software~\citep{yang2024sweagent,wang2024openhands}, while computer use agents act through browser and desktop interfaces, adapting to interface changes~\citep{zhou2023webarena,xie2024osworld}. Multiagent frameworks distribute workflows among specialized roles that communicate and share intermediate results~\citep{wu2023autogen,hong2023metagpt}. These workflows establish mechanisms for tool use, memory, feedback, and coordination. AGP studies how they shape the robot policy boundary, with agents governing physical task execution through perception, geometry, planning, and control interfaces.

\section{A General Purpose Agent as the Robot Policy}
\label{sec:method}

\subsection{Problem Formulation}
\label{sec:method:overview}

\noindent\textbf{Objective.} We consider physical manipulation tasks specified by video, image, or language instructions, or a combination of these. A robot operates in a persistent scene. Its objective is to complete the task.

\noindent\textbf{Task preparation.} When a new task is introduced, a preparation agent organizes the task instructions, reference images or demonstration videos, and robot interface documentation into a task definition. This definition specifies the goal, task constraints, completion criteria, reference materials, and tool access. The execution agent reads this definition together with the current instance information and carries out the task.

\noindent\textbf{Agent as the robot policy.} The execution agent is a general purpose agent (e.g., Codex) that serves directly as the robot policy. It receives a task specification $g$ and a documented robot interface $I$. The task specification includes the instructions collected by the preparation agent and other demonstration materials. The agent maintains a local workspace $W$ with scripts, notes, and observations collected during execution. It interprets the task, plans actions, and interacts with the robot through tools. We describe its tool selection process as
\begin{equation}
    u_k \sim \pi_{\theta}(\,\cdot \mid g, I, H_k, W_k),
    \label{eq:runtime-policy}
\end{equation}
where $u_k$ is the tool call selected at step $k$. Here, $H_k$ contains the conversation and tool results available at that step, and $W_k$ contains the current workspace files. A tool call may gather information or run a program that submits action targets to the robot controller. The model parameters $\theta$ remain fixed throughout execution.

\subsection{Agent-Robot Interface}
\label{sec:method:execution}

The robot interface $I$ supports two-way information exchange between the agent and the robot. The agent sends observation and action requests and receives camera observations, robot state, and action results. The interface is implemented as a Python command-line client and a separate server process. The agent calls the client, which exchanges requests and responses with the server through JSON files in the agent's workspace. The server communicates with the robot and cameras through a separate program that handles hardware access.

\textbf{Robot to agent.} The interface provides two sources of information: camera observations and robot state. An overhead camera provides a view of the scene, and another wrist camera provides close-up images and depth values aligned with the image pixels. Camera calibration data and camera poses allow the agent to convert these visual measurements into positions in the robot's coordinate frame. The interface also reports joint angles measured by the motors, together with the gripper's position and orientation computed from these angles through forward kinematics. After a motion, the interface reports the available updated state and errors relative to the requested target. This feedback helps the agent check whether the robot reached the intended pose. Observations are saved in the agent's workspace so it can view them or process them with programs.

\textbf{Agent to robot.} The agent sends two types of requests through the interface. \textit{Observation requests} ask the robot to capture images or report its state. \textit{Action requests} specify arm movements or gripper opening and closing. To move the arm, the agent specifies either target joint angles or a target position and orientation for the gripper. Separate gripper requests specify how far the fingers should open or close. The robot controller converts these targets into commands that drive the joints and gripper. When the agent specifies a gripper pose, the controller uses Mink~\citep{zakka2026mink} to compute target joint angles through inverse kinematics. It then generates joint trajectories within workspace and motion limits and uses motor feedback to follow them. The robot controller operates independently of agent inference, so the agent can continue reasoning while the robot moves. Each arm executes motion commands in order, while different arms can move at the same time.

\subsection{Runtime Programming and Execution}
\label{sec:method:programming}

\noindent\textbf{Runtime programming.} The agent writes programs during execution to interpret observations, compute action targets, and submit them through the robot interface. It can inspect reference videos and camera images and process saved data with Python and available libraries. For example, it can estimate an object's position from camera observations and depth measurements, compute a target gripper pose, and issue the corresponding robot commands. The supplementary material provides an analysis of recorded programs for geometric estimation, visual processing, and robot call composition.

\noindent\textbf{Feedback and adaptation.} After executing an action, the agent uses motion feedback and new observations to revise its estimates and subsequent actions. For example, an unsuccessful insertion can prompt a new alignment estimate and a modified approach. The agent determines grasp poses, action sequences, observation timing, and recovery during this cycle. It continues until it reports completion or ends the attempt. This report only terminates execution: physical success is judged by human assessment against the task-specific success criteria provided in the supplementary material using the resulting scene and recorded evidence.

\subsection{Execution Efficiency}
\label{sec:method:efficiency}

When control returns to the agent after a tool call, choosing the next operation requires another model inference round over the available context and new tool results. Frequent returns therefore add inference latency and token usage, including repeated input context across requests. We introduce \textit{AGP\_accelerate} to reduce this overhead by grouping consecutive operations and reusing geometric estimates while retaining execution feedback.

\noindent\textbf{Action batching with feedback.} To reduce inference rounds between consecutive operations, the agent submits short sequences containing measurements, arm movements, and gripper commands. The runtime executes each step using the preceding step's feedback and resolves targets through named measurements. A failed measurement (e.g., missing depth at a selected pixel) or unexpected motion outcome (e.g., a target pose mismatch due to motor error) stops the sequence and returns current observations to the agent. This allows routine operations to proceed within one inference round while preserving opportunities for correction.

\noindent\textbf{Reusable geometric estimates.} To reduce repeated geometry programming, the runtime converts selected image points into robot coordinates using calibrated depth. It evaluates candidate measurements and exposes the selected targets for subsequent actions to reuse. After approaching a target, it can refine the estimate using wrist depth. Contact detected through motor feedback or intentionally initiated by the agent invalidates affected measurements. This supports continuous execution through reusable estimates whose validity is tracked as the scene changes.

\section{Experiments}
\label{sec:experiments}

We evaluate AGP on real-world robot tasks spanning long horizon reasoning, action diversity, and object diversity under video, image, and language instructions.

\subsection{Experimental Setup}
\label{sec:exp_setup}

\subsubsection{Robot and Agent}
\label{sec:exp_platform}

\noindent\textbf{Robot Platform.} We use an I2RT YAM~\citep{i2rt_yam} arm with six revolute joints and a parallel gripper in a tabletop workspace. AGP provides the robot interface described in Section~\ref{sec:method:execution}. The towel folding task uses two arms with a common coordinate frame and coordination selected by the agent. Robot platform and control interface details are provided in the supplementary material.

\noindent\textbf{Cameras.} A wrist mounted Intel RealSense D405 provides aligned RGB and depth at $640\times360$ pixels. A fixed overhead Logitech BRIO provides rectified RGB at $1920\times1080$ pixels. Both cameras capture at 30 frames per second, and the agent requests observations on demand.

\noindent\textbf{Model Configuration.} We use Codex and Claude Code as robot policies in our experiments. Model weights remain fixed during execution. Section~\ref{sec:exp_open_models} additionally studies open models and offline trajectory distillation.

\begin{table}[t]
    \centering
    \begin{minipage}{\linewidth}
    \centering
    \caption{\textbf{Main results using GPT-6 Astra as policy.} Success is shown as successful trials over total trials ($n/N$). Time, tokens, and cost show means followed by [min, max], calculated only on successful trials. Time measures task delivery through task completion. Tokens count input and output usage, including reasoning tokens and retries. Cost uses Standard API rates with cached input pricing. Rows tagged AGP\_accelerate use high effort. Tags denote long horizon reasoning (LHR), action diversity (AD), object diversity (OD), video instruction (Vid), image instruction (Img), and language instruction (Lan).}
    \label{tab:task_comparison}
    \small
    \setlength{\tabcolsep}{7pt}
    \renewcommand{\arraystretch}{0.95}
    % Successful trial usage audit is analysis/successful_trial_metrics/per_run.csv.
    % Trial selection, timing, and usage are in analysis/main_task_timing/.
    % Completion time spans task delivery through completion of task execution.
    \newcommand{\meanrange}[3]{#1\,{\scriptsize\textcolor{black!65}{[#2, #3]}}}
    \resizebox{\linewidth}{!}{%
    \begin{tabular}{@{}lcccccc@{}}
        \toprule
        Task & Capability & Instruction & Success & Time $\downarrow$ & Tokens $\downarrow$ & Cost $\downarrow$ \\
        & & & $(n/N)$ & (min) & (million) & (USD) \\
        \midrule
        % Assembly sources and accounting are in analysis/four_pair_assembly/per_run.csv.
        Four pair assembly & \captag{capReasoning}{LHR} & \captag{capVisual}{Vid} & 8/10 & \meanrange{37.2}{16.0}{59.1} & \meanrange{13.02}{3.96}{27.12} & \meanrange{16.62}{5.55}{31.50} \\
        \quad \captag[\width]{capAccelerate}{AGP\_accelerate} & &  & 5/5 & \meanrange{7.6}{5.3}{8.7} & \meanrange{2.47}{1.54}{3.51} & \meanrange{3.82}{2.58}{5.08} \\
        \cmidrule[0.25pt](lr){1-7}
        \begin{tabular}[c]{@{}l@{}}Block construction\\[0pt]\quad \textit{Pyramid}\end{tabular} & \captag{capReasoning}{LHR} & \captag{capVisual}{Img} & 10/10 & \meanrange{21.6}{14.9}{32.2} & \meanrange{9.24}{6.27}{12.94} & \meanrange{11.69}{8.01}{15.28} \\
        \qquad \captag[\width]{capAccelerate}{AGP\_accelerate} & &  & 5/5 & \meanrange{10.0}{7.2}{13.8} & \meanrange{2.68}{1.50}{4.80} & \meanrange{4.19}{2.63}{6.95} \\
        \quad \textit{Two towers} & & & 10/10 & \meanrange{20.6}{14.8}{29.2} & \meanrange{7.94}{4.94}{14.39} & \meanrange{9.92}{6.27}{16.85} \\
        \qquad \captag[\width]{capAccelerate}{AGP\_accelerate} & &  & 5/5 & \meanrange{6.6}{5.3}{8.5} & \meanrange{1.64}{1.03}{2.21} & \meanrange{2.82}{2.03}{3.61} \\
        \quad \textit{Six block tower} & & & 9/10 & \meanrange{28.2}{11.6}{55.0} & \meanrange{11.57}{4.50}{19.85} & \meanrange{14.93}{6.50}{26.78} \\
        \cmidrule[0.25pt](lr){1-7}
        Dice flipping & \captag{capAction}{AD} & \captag{capText}{Lan} & 10/10 & \meanrange{37.9}{22.6}{56.0} & \meanrange{17.08}{10.16}{25.00} & \meanrange{21.07}{12.57}{31.20} \\
        \quad \captag[\width]{capAccelerate}{AGP\_accelerate} & & & 5/5 & \meanrange{17.1}{9.6}{22.1} & \meanrange{5.22}{1.93}{8.54} & \meanrange{7.52}{3.41}{11.47} \\
        \cmidrule[0.25pt](lr){1-7}
        % Selected sessions and accounting are in analysis/throwing/per_run.csv.
        Targeted throwing & \captag{capAction}{AD} & \captag{capText}{Lan} & 9/10 & \meanrange{11.1}{5.9}{21.4} & \meanrange{5.75}{2.58}{14.15} & \meanrange{8.24}{4.10}{18.97} \\
        \quad \captag[\width]{capAccelerate}{AGP\_accelerate} & &  & 4/5 & \meanrange{3.6}{2.7}{5.1} & \meanrange{0.75}{0.61}{1.00} & \meanrange{1.69}{1.42}{2.00} \\
        \cmidrule[0.25pt](lr){1-7}
        % Sources are paper_runs/towel{,2}_full_high_yieldbatch_v2_kn_dual/results.csv.
        % Timing and usage audit is analysis/towel_folding/per_run.csv.
        % Success counts follow user review of final images on September 10, 2026.
        \begin{tabular}[c]{@{}l@{}}Towel folding\\[0pt]\quad \textit{Sequential}\end{tabular} & \captag{capObject}{OD} & \captag{capVisual}{Vid} & 5/5 & \meanrange{50.8}{33.8}{63.3} & \meanrange{18.12}{10.13}{25.26} & \meanrange{24.14}{14.16}{30.90} \\
        \quad \textit{Simultaneous} & & & 3/5 & \meanrange{21.4}{15.7}{31.2} & \meanrange{5.06}{4.27}{5.68} & \meanrange{7.46}{6.16}{8.67} \\
        \bottomrule
    \end{tabular}
    }
    \end{minipage}
\end{table}

\subsubsection{Tasks and Evaluation}
\label{sec:exp_evaluation}

\noindent\textbf{Task Design.} We find that conventional ``pick and place'' tasks, such as placing an object into a bowl, pose little challenge for AGP, motivating more demanding tasks to evaluate its capabilities. Therefore, we use five different task groups. Task definitions, success criteria, and execution examples are provided in the supplementary material.
\begingroup
\setlength{\topsep}{1pt}
\setlength{\partopsep}{0pt}
\setlength{\leftmargini}{2.5em}
\begin{itemize}
    \item \textbf{Four pair assembly.} The robot assembles eight parts into four pairs. It follows a human demonstration video and must identify the pairings and order the operations while preserving earlier progress. This task tests \emph{long horizon reasoning} under video instructions. The parts are adapted from the AutoMate dataset~\citep{tang2024automate}.
    \item \textbf{Block construction.} The robot builds a block structure from a goal image. The targets comprise a pyramid, two towers, and a six-block tower. This task tests \emph{long horizon reasoning} under image instructions. It requires interpreting spatial relationships and ordering placements while preserving earlier progress.
    \item \textbf{Dice flipping.} The robot reorients six randomly placed dice. All six must show the number specified by the language instruction on their upward faces. This task tests \emph{action diversity}: the agent must choose grasps and rotations based on each die's current orientation and account for how it settles after release.
    \item \textbf{Targeted throwing.} The robot throws an object toward a target. The target is specified through language instructions. This task tests \emph{action diversity} because the robot must release the object during a circular swing. The agent must coordinate the swing with gripper opening to send the object toward the target.
    \item \textbf{Towel folding.} The robot folds a towel using two arms. It follows video instructions and tests \emph{object diversity}: it must coordinate both arms to handle a soft and deformable object.
\end{itemize}
\endgroup

\noindent\textbf{Trial Protocol \& Eval Metrics.} For the original and accelerated AGP evaluations in Table~\ref{tab:task_comparison}, each trial starts with a fresh execution workspace containing the task definition, task reference images or demonstration videos, and robot interface documentation. For all tasks, initial object positions are randomized within the robot's reachable workspace in each trial. Each trial is allotted a maximum of 120 minutes. For each agent, we vary the task order to reduce order bias. The agent may choose to autonomously recover from errors. We manually assess success against the task-specific success criteria using the physical outcomes and the execution process. We primarily evaluate AGP by its success rate. We also report completion time, token usage, and model inference cost.

\subsection{Experimental Results}
\label{sec:exp_results}

\subsubsection{Main Results}
\label{sec:exp_main_results}

Table~\ref{tab:task_comparison} reports task success and execution efficiency for AGP and AGP\_accelerate using the same GPT-6 Astra model, high thinking effort, and default service tier. The original AGP evaluation covers eight task configurations spanning long horizon reasoning, action diversity, and object diversity. AGP\_accelerate is evaluated on five of these configurations to assess the efficiency gains from Section~\ref{sec:method:efficiency}.
\begingroup
\setlength{\topsep}{1pt}
\setlength{\partopsep}{0pt}
\setlength{\leftmargini}{2.5em}
\begin{itemize}
    \item \textbf{Strong zero-shot performance.} AGP achieves success rates of at least 80\% in seven of eight task configurations and succeeds in every trial in four configurations. These results support its effectiveness for zero-shot manipulation across the evaluated tasks.
    \item \textbf{Improved execution efficiency.} AGP\_accelerate substantially reduces execution time and inference overhead. Across all five evaluated task configurations, mean completion time falls by more than half relative to the original AGP, while mean token usage and inference cost each decrease by over 60\%.
    \item \textbf{Task complexity and execution overhead.} More complex tasks generally take longer and incur higher inference costs. For example, building the six-block tower takes more time and costs more than building the pyramid or the two towers.
    \item \textbf{Challenges in deformable object manipulation.} Sequential towel folding succeeds in all five trials but takes 50.8 minutes and costs USD 24.14 on average. Both averages are the highest among the evaluated tasks. Simultaneous folding follows a demonstration in which both short ends fold inward together. It has the lowest observed success rate at 3/5, highlighting the difficulty of coordinating movements when manipulating deformable material.
\end{itemize}
\endgroup

\subsubsection{Agent and Thinking Effort Comparison}
\label{sec:exp_models}

\newsavebox{\modelcomparisonbox}
\sbox{\modelcomparisonbox}{%
    \small
    \setlength{\tabcolsep}{3pt}
    \renewcommand{\arraystretch}{1.05}
    % Means above [min, max] over successful trials; macro local to this table.
    \newcommand{\meanrange}[3]{\begin{tabular}[c]{@{}c@{}}#1\\[-3pt]{\scriptsize\textcolor{black!65}{[#2, #3]}}\end{tabular}}
    \begin{tabular}{@{}lcccc@{}}
        \toprule
        Model & Success & Time & Tokens & Cost \\
        & $(n/N)$ & (min) & (million) & (USD) \\
        \midrule
        \multicolumn{5}{@{}l}{\textit{Codex}} \\
        GPT-6 Astra & & & & \\
        % Astra low: 5 trials. Source selection and accounting are in paper_runs/table2_metrics.csv.
        \quad \textit{Low effort} & 5/5 & \meanrange{9.9}{6.0}{19.0} & \meanrange{3.69}{1.44}{7.72} & \meanrange{4.79}{2.15}{9.56} \\
        % Astra medium: 5 trials. Source selection and accounting are in paper_runs/table2_metrics.csv.
        \quad \textit{Medium effort} & 5/5 & \meanrange{9.2}{6.3}{15.4} & \meanrange{3.03}{1.84}{5.80} & \meanrange{4.09}{2.76}{7.48} \\
        % Astra high: 5 trials. Source selection and accounting are in paper_runs/table2_metrics.csv.
        \quad \textit{High effort} & 5/5 & \meanrange{9.2}{7.9}{10.6} & \meanrange{3.30}{2.58}{4.02} & \meanrange{4.47}{3.72}{5.19} \\
        GPT-5.6 Sol & 5/5 & \meanrange{14.2}{9.1}{19.3} & \meanrange{7.13}{4.12}{9.72} & \meanrange{3.94}{2.40}{5.30} \\
        % Terra trial accounting: gap_yam_real/free_agent/paper_runs/table2_metrics.csv (tools/paper_table2_metrics.py, same boundary rule as the Sol rows).
        GPT-5.6 Terra & 1/5 & \meanrange{19.0}{19.0}{19.0} & \meanrange{6.28}{6.28}{6.28} & \meanrange{1.98}{1.98}{1.98} \\
        GPT-5.6 Luna & 0/5 & -- & -- & -- \\
        \noalign{\vskip 1pt}
        \multicolumn{5}{@{}c@{}}{\dotfill} \\[1pt]
        \multicolumn{5}{@{}l}{\textit{Claude Code}} \\
        % Opus 5 via Claude Code (print mode): 5 trials. Source selection and accounting are in paper_runs/table2_metrics.csv.
        Claude Opus 5 & 5/5 & \meanrange{22.2}{14.4}{31.9} & \meanrange{12.40}{9.05}{16.73} & \meanrange{9.75}{6.94}{13.03} \\
        % Fable successful trial accounting is in analysis/fable_table2/per_run.csv.
        Claude Fable 5.1 & 3/5 & \meanrange{27.7}{18.3}{39.8} & \meanrange{9.27}{6.87}{11.10} & \meanrange{11.30}{8.33}{14.36} \\
        \bottomrule
    \end{tabular}%
}

\WFclear
\Needspace{300pt}
\begin{wraptable}{r}{0.55\textwidth}
    \captionsetup{font=small}
    \centering
    \caption{\textbf{Agent and thinking effort comparison on a two-pair assembly task.} Time, tokens, and cost show means with [min, max] below, calculated only on successful trials. Models other than GPT-6 Astra use high thinking effort.}
    % Sol trial accounting is in analysis/two_pair_assembly/per_run.csv.
    \label{tab:model_effort_assembly}
    \resizebox{\linewidth}{!}{\usebox{\modelcomparisonbox}}
\end{wraptable}

Table~\ref{tab:model_effort_assembly} compares agent and model configurations on a two-pair assembly task based on video demonstration. The task pairs a hexagonal ring with a short post and a circular sleeve with a stepped cylinder. We compare GPT-6 Astra at low, medium, and high thinking effort and evaluate additional models listed in the table. All configurations share task references and robot functions.

AGP completes the assembly task with both Codex and Claude Code, with model choice affecting success and efficiency. Astra succeeds in all five trials at each thinking effort level, with similar mean completion times of 9.2 to 9.9 minutes. Sol also succeeds in all five trials at lower inference cost than Astra, but takes longer. Terra succeeds in one of five trials and Luna in none. Within Claude Code, Opus succeeds in all five trials and Fable in three. On successful trials, Fable uses fewer tokens than Opus but takes longer and costs more.

\WFclear
\subsubsection{Baseline Comparison}
\label{sec:exp_baselines}

\newsavebox{\baselinecomparisonbox}
\sbox{\baselinecomparisonbox}{%
    \small
    \setlength{\tabcolsep}{3pt}
    \renewcommand{\arraystretch}{1.05}
    % Audit and exact source selection are in analysis/baseline_metrics/audit.json.
    % Real counts follow author confirmation on 2026-09-25.
    % Simulation sources and author-confirmed GaP counts are in analysis/baseline_simulation_success.json.
    \begin{tabular}{@{}lcccc@{}}
        \toprule
        Method & \multicolumn{2}{c}{Success $(n/N)$ $\uparrow$} & Time $\downarrow$ & Cost $\downarrow$ \\
        & Sim. & Real & (min) & (USD) \\
        \midrule
        \multicolumn{5}{@{}l}{\textit{Two pair assembly}} \\
        GaP & \textcolor{red!75!black}{8/25} & \textcolor{red!75!black}{0/5} & -- & -- \\
        ASPIRE & \textcolor{green!45!black}{25/25} & \textcolor{red!75!black}{2/5} & 2.9 / 64.8 & 0.00 / 18.90 \\
        AGP & -- & \textcolor{green!45!black}{5/5} & 9.2 & 4.47 \\
        \midrule
        \multicolumn{5}{@{}l}{\textit{Four pair assembly}} \\
        GaP & \textcolor{red!75!black}{1/25} & \textcolor{red!75!black}{0/10} & -- & -- \\
        ASPIRE & \textcolor{green!45!black}{20/25} & \textcolor{red!75!black}{0/10} & -- & -- \\
        AGP & -- & \textcolor{green!45!black}{8/10} & 37.2 & 16.62 \\
        \midrule
        \multicolumn{5}{@{}l}{\textit{Pyramid construction}} \\
        GaP & \textcolor{red!75!black}{3/25} & \textcolor{red!75!black}{0/10} & -- & -- \\
        ASPIRE & \textcolor{green!45!black}{25/25} & \textcolor{red!75!black}{1/10} & 9.2 / 280.9 & 0.00 / 78.13 \\
        AGP & -- & \textcolor{green!45!black}{10/10} & 21.6 & 11.69 \\
        \bottomrule
    \end{tabular}%
}

\WFclear
\begin{wrapfigure}{R}{0.49\textwidth}
    \captionsetup{font=small}
    \centering
    \includegraphics[width=\linewidth]{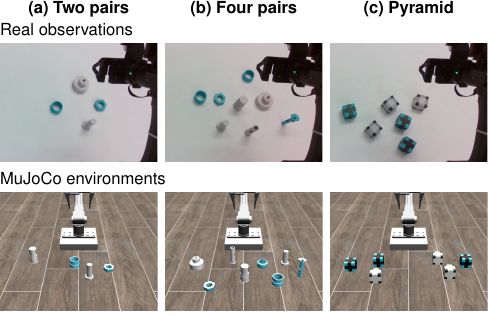}
    \caption{\textbf{Real and simulated task environments.} The real robot task scenes (top) are reproduced in MuJoCo at matching physical scales (bottom) to support simulation refinement for GaP and ASPIRE.}
    \label{fig:baseline_environments}
\end{wrapfigure}

We compare with two representative agentic robotics methods, Graph-as-Policy (GaP)~\citep{chen2026gap} and ASPIRE~\citep{lu2026aspire}. GaP extends Code as Policies~\citep{liang2022code} with multiple agents that organize perception, planning, and control skills into a computation graph, then refine its structure and parameters through simulation feedback. ASPIRE uses execution traces for debugging, evolutionary search over programs, and a library of reusable skills. We evaluate AGP against these two methods on two pair assembly, four pair assembly, and pyramid construction.

\noindent\textbf{Baseline preparation and evaluation.} We adapt the official implementations of GaP and ASPIRE to the YAM platform. Both baselines use the same robot SDK as AGP for physical execution. All three methods use GPT-6 Astra with high thinking effort. We construct MuJoCo environments with the YAM robot model, assembly objects at their printed scales, and 50\,mm cubes (Figure~\ref{fig:baseline_environments}). We prepare both baselines in these environments. For GaP, generated graphs are evaluated in simulation and then executed on the physical robot. Following the ASPIRE procedure, we test and refine the program constructed in simulation on the physical robot to reduce the gap between simulation and physical execution. ASPIRE uses real camera observations, execution traces, and physical outcomes to revise its perception and manipulation code. The refined program is then frozen for subsequent evaluation. AGP executes zero-shot on the physical robot.

\newpage

\noindent\textbf{Time and Cost Measurement.} For ASPIRE, Table~\ref{tab:baseline_comparison} reports real only / sim + real time and cost. Real only covers successful program execution, while sim + real adds the full shared construction and refinement overhead in simulation and on the real robot to each successful run, without amortization. AGP time covers task delivery through task completion, including reasoning, programming, execution, and recovery.

\noindent\textbf{Task Success.} GaP fails to complete the evaluated real robot tasks (Table~\ref{tab:baseline_comparison}). Its simulation failures include cases where the inverse kinematics (IK) solver cannot find a feasible solution or visual recognition fails. These errors terminate execution in the evaluated graphs, whose recovery logic is limited.

\WFclear
\Needspace{290pt}
\begin{wraptable}{r}{0.55\textwidth}
    \captionsetup{font=small}
    \centering
    \caption{\textbf{Baseline comparison.} Sim. reports preparation-stage simulation validation, and Real reports physical trials. AGP is evaluated directly on the robot. Time and cost average successful real runs. ASPIRE reports real only / sim + real, with full construction and refinement overhead in the latter. Execution makes no coding model calls, so its token cost is zero. Green/red marks success above/below 50\%.}
    \label{tab:baseline_comparison}
    \resizebox{\linewidth}{!}{\usebox{\baselinecomparisonbox}}
\end{wraptable}

ASPIRE achieves successful physical executions during refinement, yet shows limited success when the resulting programs are frozen for evaluation. The fixed program structure may limit adaptation to changes in object placement or unexpected action outcomes. AGP succeeds on most trials across all three tasks, using fresh observations to revise object estimates, adjust grasp and insertion strategies, and recover from unsuccessful attempts.

\noindent\textbf{Efficiency.} ASPIRE's time and inference cost are concentrated in program construction and refinement. Once prepared, the programs can be reused without further coding model inference cost. In our evaluation, the low task success rate limits the practical value of this program reuse. When the full preparation overhead is charged to each successful run, AGP takes less time and incurs lower inference cost on two pair assembly and pyramid construction.

\WFclear
\subsection{Experience Refinement and Transfer}
\label{sec:exp_efficiency}

\begin{figure}[t]
    \centering
    \includegraphics[width=\linewidth]{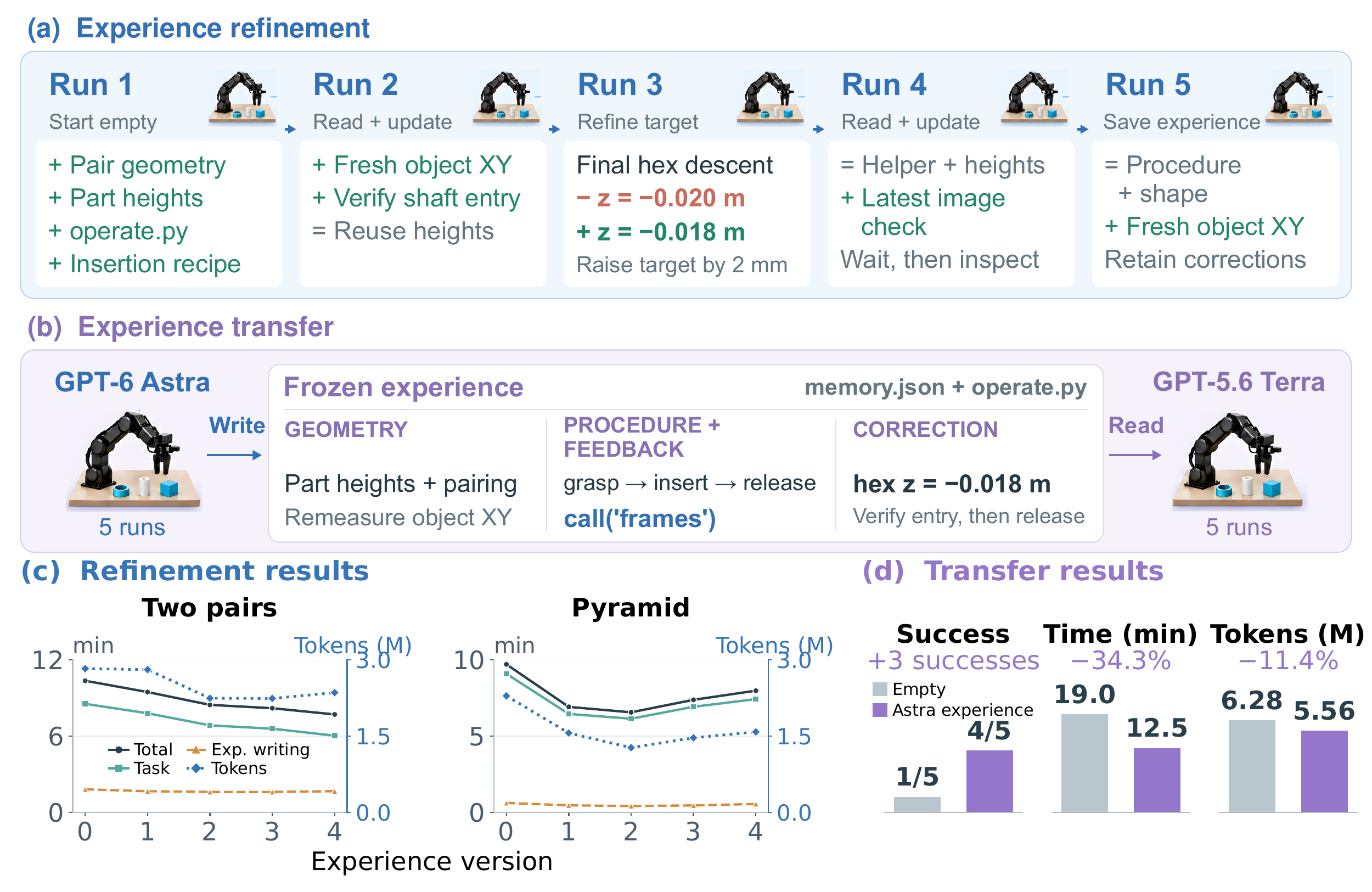}
    \caption{\textbf{Experience refinement and transfer.} (a) Experience updates across runs. (b) Transfer of frozen experience from Astra to Terra. (c) Time on the left axes and tokens on the right axes share the experience version axis. Version 0 uses empty experience. Time scales differ across tasks. (d) Transfer results on two pair assembly.}
    \label{fig:experience_refinement_transfer}
\end{figure}

We study \emph{experience refinement} to improve execution efficiency and \emph{experience transfer} to improve a weaker model's task success rate (Figure~\ref{fig:experience_refinement_transfer}). In these studies, experience is provided through documents saved during earlier runs of the same task. Refinement starts with five executions that update saved experience after each run, yielding five starting versions from empty experience through four updates. For both tasks, we evaluate each version in three trials initialized from the same saved documents and report mean runtime and token usage. Transfer gives a weaker model the experience recorded by a stronger model.

\phantomsection\label{sec:exp_experience_accumulation}
\noindent\textbf{Experience Refinement.} We test GPT-6 Astra on two pair assembly and pyramid construction. Saved documents contain successful grasp and insertion settings, corrections, and reusable programs. Later runs update object coordinates while reusing this knowledge (Figure~\ref{fig:experience_refinement_transfer}(a)). Figure~\ref{fig:experience_refinement_transfer}(c) compares runtime and token usage across experience versions. Runtime includes experience writing. Despite fluctuations across subsequent runs, all four versions with prior experience have lower mean total runtime and token usage than empty experience. For pyramid construction, mean total runtime and task token usage decrease by 17.8\% and 30.9\%, respectively, between the first and fifth experience versions.

\phantomsection\label{sec:exp_experience_transfer}
\noindent\textbf{Experience Transfer.} We transfer experience from Astra's five two-pair assembly runs to GPT-5.6 Terra. Terra performs five trials with Astra's saved experience and five without it. Every trial with experience reads the same documents without updating them. Figure~\ref{fig:experience_refinement_transfer}(d) shows higher task success with transferred experience. Among successful trials, mean completion time and token usage decrease by 34.3\% and 11.4\%, respectively.

\subsection{Case Study of Failure Recovery}
\label{sec:exp_failure_recovery}

\begin{figure}[t]
    \centering
    \includegraphics[width=0.98\linewidth]{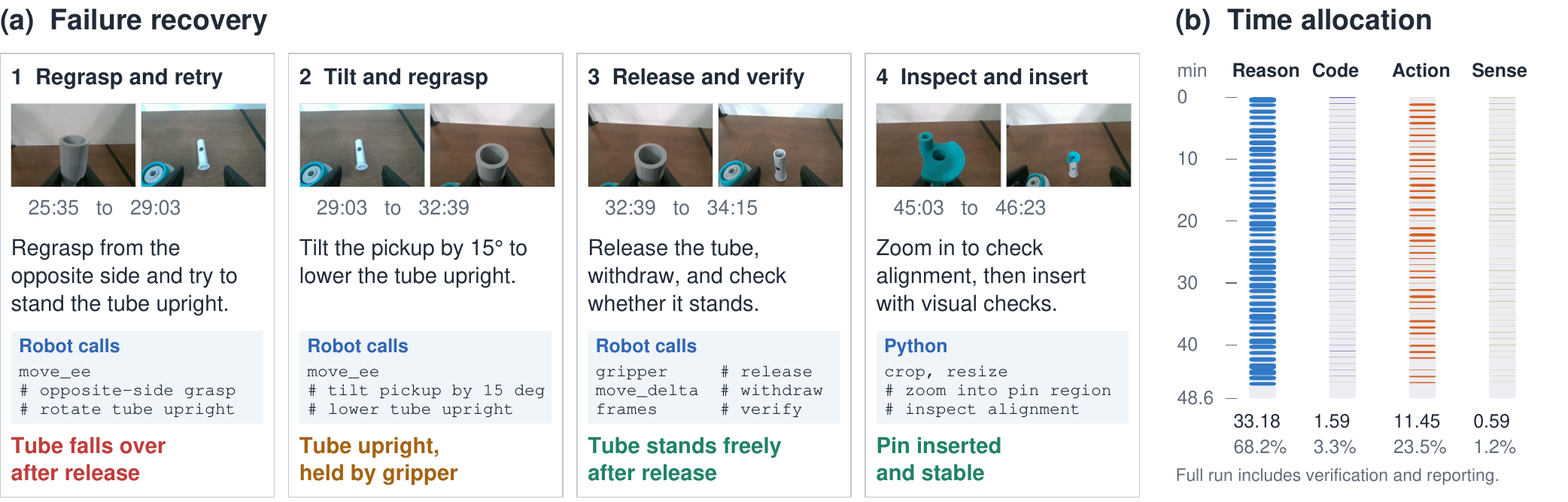}
    \caption{\textbf{Failure recovery during four pair assembly.} (a) Visual feedback guides a tilted pickup after the tube falls over, followed by alignment checks during insertion. Timestamps show elapsed time. (b) Vertical timelines show reasoning (Reason), program execution (Code), robot actions (Action), and camera and state calls (Sense), with elapsed time increasing downward. Values below each timeline give minutes and the share of the full run.}
    \label{fig:failure_recovery}
\end{figure}

In a four pair assembly run, AGP revises its grasp after observing that the tube falls over upon release (Figure~\ref{fig:failure_recovery}(a)). It introduces a $15^{\circ}$ pickup tilt, lowers the tube upright with the wrist angled near the table, and withdraws the gripper to verify stability. For insertion, it enlarges a camera image to inspect pin alignment, lowers the pin gradually, and checks fresh images until the released assembly is stable. Figure~\ref{fig:failure_recovery}(b) breaks down the 48.6 minute run. Planning and reasoning account for 68.2\% of runtime, and robot actions for 23.5\%.

% Slightly tighter spacing for the two compact result tables.
\begingroup
\setlength{\intextsep}{6pt plus 1pt minus 1pt}
\captionsetup[table]{skip=6pt}
\subsection{Open Weight Models}
\label{sec:exp_open_models}

We study whether locally hosted open weight models can serve as AGP's runtime policy. We evaluate in simulation on test sets of VLA-Arena~\citep{zhang2025vlaarena} and RoboTwin 2.0~\citep{chen2025robotwin2} benchmarks, using three models as the AGP policy: Qwen3.6-35B-A3B, Qwen3.6 after our SFT, and Qwen3.8-Flash-Next.

\noindent\textbf{Trajectory distillation.} For our SFT, we collect 5,658 successful trajectories from simulated AGP executions with Astra as the policy, and distill Qwen3.6 from them. Training data covers the training splits of LIBERO family, RoboTwin, VLA-Arena, and RoboCasa, without any task overlap with the test sets. We update all language model parameters and freeze the vision encoder. Training configurations are provided in the supplementary material.

\WFclear
\Needspace{135pt}
\begin{wraptable}{r}{0.55\textwidth}
    \captionsetup{font=small}
    \centering
    \caption{\textbf{AGP policies on two simulation benchmarks.}}
    \label{tab:open_model_simulation}
    \small
    \setlength{\tabcolsep}{3pt}
    \begin{tabular*}{\linewidth}{@{\extracolsep{\fill}}lcccc@{}}
        \toprule
        Benchmark & \begin{tabular}[c]{@{}c@{}}GPT-6\\Astra\end{tabular}
        & \begin{tabular}[c]{@{}c@{}}Qwen3.6\\35B-A3B\\Original\end{tabular}
        & \begin{tabular}[c]{@{}c@{}}Qwen3.6\\35B-A3B\\SFT\end{tabular}
        & \begin{tabular}[c]{@{}c@{}}Qwen3.8\\Flash-Next\\Original\end{tabular} \\
        \midrule
        VLA-Arena & 70.6\% & 0.8\% & 57.5\% & 33.7\% \\
        \addlinespace[3pt]
        RoboTwin & 82.5\% & 8.3\% & 62.3\% & 48.0\% \\
        \bottomrule
    \end{tabular*}
\end{wraptable}

\noindent\textbf{Simulation results.} As shown in Table~\ref{tab:open_model_simulation}, Qwen3.6 SFT achieves 57.5\% on VLA-Arena and 62.3\% on RoboTwin, despite being largely unable to perform these tasks before training. This shows that, given training data of decent quality, small open weight models can learn the behaviors required of a robot policy and serve as AGP's backbone model. Future AGP systems therefore need not depend on the strongest closed source models. Moreover, Qwen3.8-Flash-Next completes tasks through AGP without our SFT, suggesting that stronger open weight models can control robots within our framework and substantially reduce inference costs.

\WFclear
\Needspace{100pt}
\begin{wraptable}{r}{0.52\textwidth}
    \captionsetup{font=small}
    \centering
    \caption{\textbf{Real-world evaluation on four object packing.}}
    \label{tab:open_model_real}
    \small
    \setlength{\tabcolsep}{3pt}
    \begin{tabular}{@{}lccc@{}}
        \toprule
        Model & Placement & Time & API cost \\
        & $(n/N)$ & (s) & (USD) \\
        \midrule
        GPT-6 Astra & 5/5 & 413.4 & 2.49 \\
        Qwen3.8-Flash-Next & 5/5 & 582.2 & -- \\
        \bottomrule
    \end{tabular}
\end{wraptable}

\noindent\textbf{Real-world results.} On a real-world task of packing four objects into a box, locally served Qwen3.8-Flash-Next and Astra both complete the task with a 100\% success rate (Table~\ref{tab:open_model_real}), although Qwen3.8-Flash-Next takes longer.

% \subsection{Applications}
% \label{sec:exp_applications}
%
% \paragraph{Data collection for policy learning.}
% AGP offers a potential way to collect robot experience through repeated task execution over extended periods. Even when individual tasks take substantial time, continued operation could accumulate trajectories for training other policy models. We study this application by recording task instructions, camera video, timestamped observations, commanded actions, measured joint and gripper states, and episode outcomes. Synchronizing these records associates visual observations with actions and resulting motion. Successful trajectories can provide demonstrations for imitation learning, while failed attempts and recovery segments can support studies of robustness.
%
% The collection study reports usable trajectories per hour, successful trajectory yield, task coverage, and human time spent on scene resets and interventions. These measures assess the data obtained over sustained operation, including time between trials. To evaluate downstream utility, we train a policy on curated AGP trajectories and test it on held out task configurations. The study specifies the policy, dataset size, and filtering criteria, and compares its performance with a matched training setup that excludes the collected data.

\WFclear
\endgroup

\section{Conclusion}
\label{sec:conclusion}

We introduce AGP, which uses a general purpose agent as the robot policy through runtime programming and physical feedback. In real robot experiments, AGP achieves success rates of at least 80\% in seven of eight task configurations, and significantly outperforms previous agentic robot systems. Tests with Codex and Claude Code show successful task execution across multiple models. Reusing experience saved in documents improves execution efficiency, and transferring this experience to a weaker model improves task success while reducing time and token usage on successful trials.

\section*{LLM Usage Statement}

The research uses Codex and Claude Code as robot policies and for agentic programming workflows. During manuscript preparation, the authors used generative AI tools to assist with drafting and language editing, including refining sentence structure and improving clarity and readability. The authors are responsible for reviewing and revising the resulting text to ensure that it accurately represents the research. The authors take full responsibility for the content of the manuscript, including the methodology, experimental results, interpretations, and conclusions.

\clearpage
\bibliographystyle{iclr2027_conference}
\bibliography{references}
\clearpage
\appendix
\pdfbookmark[0]{Supplementary Material}{supplement}
\section{Implementation Details}
\label{sec:appendix}

\subsection{Robot Platform and Control Interface}
\label{sec:appendix_platform}

\begin{figure}[!htbp]
    \centering
    \includegraphics[width=0.9\linewidth,height=0.38\textheight,keepaspectratio]{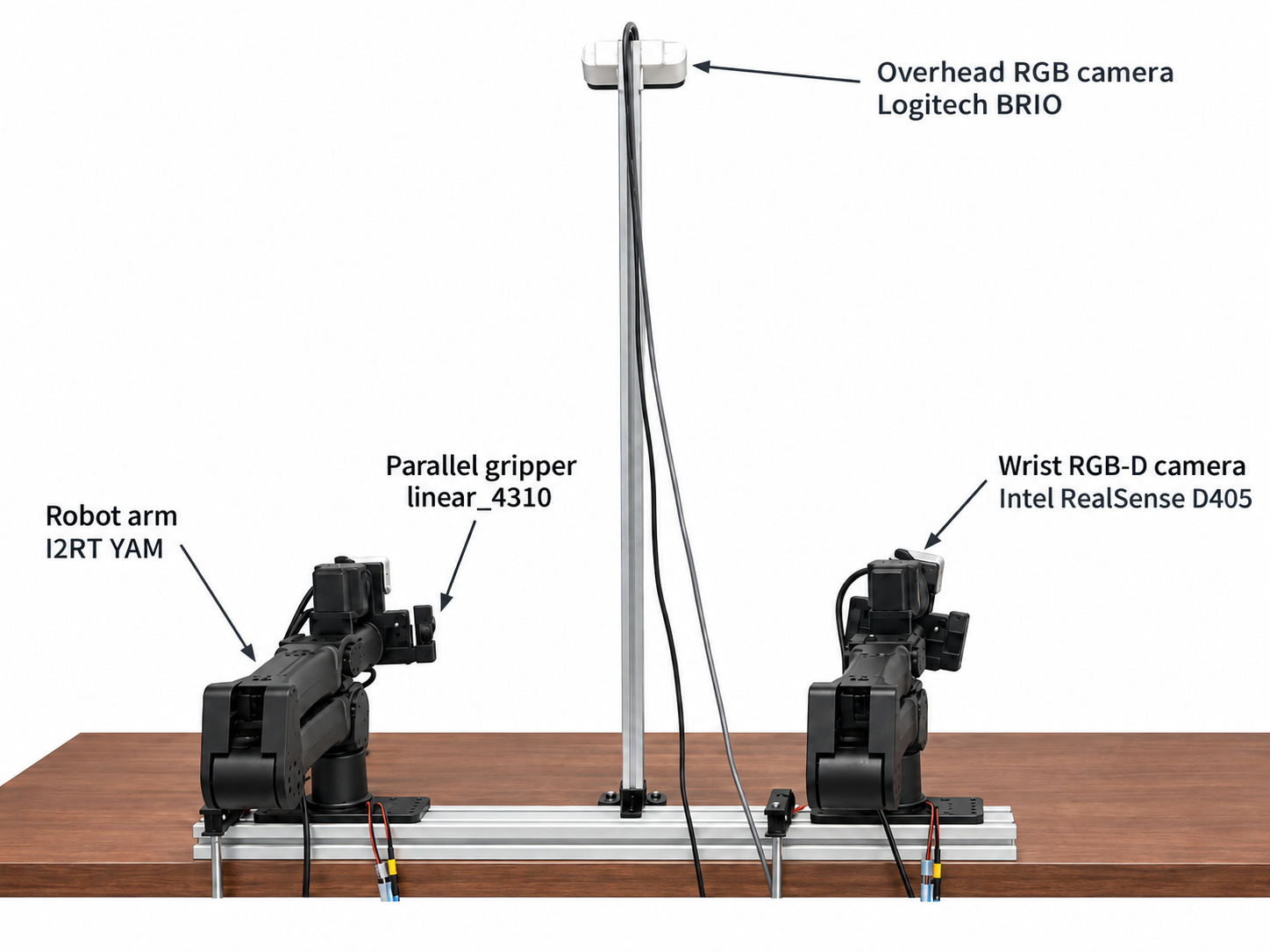}
    \caption{\textbf{Robot platform and camera placement.} The setup contains two I2RT YAM arms mounted on a shared tabletop rail.}
    \label{fig:robot_platform}
\end{figure}

Figure~\ref{fig:robot_platform} shows the platform described in Section~\ref{sec:exp_platform}.

\paragraph{Robot platform.}
The YAM arm uses a \textnormal{linear\_4310} parallel gripper with a nominal maximum inner jaw opening of 95.5\,mm. The \textnormal{i2rt} Python library supplies robot models, motor communication, and gravity compensation over a 1\,Mbit/s CAN connection. Bimanual Cartesian poses share a coordinate frame through a calibrated rigid transform.

\paragraph{Observations and geometric grounding.}
Captures from the cameras specified in Section~\ref{sec:exp_platform} include calibration, timestamps, and robot state. Wrist depth maps image pixels into the arm base frame. Overhead pixels are located by intersecting camera rays with a plane at an estimated height. The agent can request additional views to refine these estimates. Camera stream rates are independent of the agent's decision rate.

\paragraph{Action interface.}
The robot interface provides seven commands. \texttt{state} returns joint positions, the end effector pose, gripper opening fraction and width, and robot health information. \texttt{status} reports the operating status of the robot interface and coordinate frame information. \texttt{help} lists the available commands and their arguments. \texttt{frames} captures images from the requested cameras, optionally saves wrist depth, and returns paths to the saved observations and calibration data.

\texttt{move\_ee} requests a target end effector position and quaternion orientation in the shared coordinate frame and returns the achieved pose and target error, or failure information. \texttt{move\_joints} requests target joint angles in radians and returns the resulting joint positions and end effector pose. \texttt{gripper} requests opening, closing, or a normalized opening fraction between 0 and 1 and returns the resulting opening fraction and width. 
% The agent computes geometry from saved images, depth, and calibration in local programs. Cartesian motion invokes the underlying inverse kinematics and trajectory execution software. The session server enforces a radial target envelope of 0.12 to 0.65\,m, a grasp point height envelope of $-0.050$ to 0.60\,m, and a maximum Cartesian step of 0.25\,m.

% \subsection{Agent Configuration and Execution Conditions}
% \label{sec:appendix_agent}

% The execution efficiency instructions in Section~\ref{sec:method:efficiency} are held fixed within each condition. Agent traces record waiting intervals, intermediate status checks, and operations grouped within each tool call.

% \subsection{Model and Reasoning Effort Configurations}
% \label{sec:appendix_models}
% The configurations are listed in Table~\ref{tab:model_effort_assembly}. The Astra effort sweep holds the model version, agent environment, task inputs, and robot interface fixed. Comparisons across Codex and Claude Code evaluate complete agent systems under shared scenes and references. Claude Code model and effort settings use its native controls~\citep{anthropic_model_config}. Claude Opus 5 uses identifier \textnormal{claude-opus-5}. Each trial records the requested and resolved model identifiers and effective effort setting.

\section{Tasks and Evaluation Protocol}
\label{sec:appendix_evaluation}

\subsection{Task Definitions and Success Criteria}
\label{sec:appendix_tasks}

Table~\ref{tab:task_suite} defines the tasks, target capabilities, instruction types, and success criteria used in the main evaluation.

\begin{table}[t]
    \centering
    \caption{\textbf{Task definitions and success criteria.} Success or failure of every trial is determined by human review.}
    \label{tab:task_suite}
    \small
    \setlength{\tabcolsep}{3pt}
    \renewcommand{\arraystretch}{1.15}
    \begin{tabular*}{\linewidth}{@{\extracolsep{\fill}}p{0.15\linewidth}p{0.17\linewidth}p{0.15\linewidth}p{0.45\linewidth}@{}}
        \toprule
        Task & Capability & Instruction type & Task requirements and success criteria \\
        \midrule
        Four pair assembly & Long horizon reasoning & Video & Assemble eight parts into four pairs in the demonstrated order, inferring mating relationships and insertion motions from the video and current scene. Success requires all four pairs to remain assembled, preserving completed pairs throughout subsequent operations. \\
        \addlinespace[6pt]
        Block construction & Long horizon reasoning & Image & Arrange six cubes into a pyramid, two towers, or a six block tower matching the goal image. Infer object identities, orientations, and support relations, and choose a placement order that maintains intermediate stability. Success requires the complete target structure. \\
        \addlinespace[6pt]
        Dice flipping & Action diversity & Language & Reorient six randomly placed dice to the requested upward face through grasp selection and rotation. Success requires all six dice to show the requested number after placement. \\
        \addlinespace[6pt]
        Targeted throwing & Action diversity & Language & Coordinate a circular joint swing with gripper release to throw an object toward the specified target. Success requires the object to separate from the moving hand, travel through the air, and come to rest within the target region. \\
        \addlinespace[6pt]
        Towel folding & Object diversity & Video & Use both arms to reproduce the demonstrated folds. Fold the near long edge onto the far edge, then fold both short ends inward, either right then left (Sequential) or beginning together (Simultaneous). Success requires both inward folds with aligned edges. \\
        \bottomrule
    \end{tabular*}
\end{table}

\paragraph{Assembly from video.}
The AutoMate parts~\citep{tang2024automate} form a ridged sleeve and broad round base, a hexagonal nut and short post, a smooth sleeve and stepped cylinder, and a long pin and hollow tube, in demonstration order. All methods receive the same human video, extracted frames, and timestamped textual guide. Figure~\ref{fig:assembly_demo_execution} shows an execution.

\paragraph{Assembly part dimensions.}
Table~\ref{tab:assembly_print_dimensions} reports the nominal printing dimensions and mating clearances of the AutoMate meshes. Successful assembly with submillimeter nominal radial clearances illustrates precise alignment (Figure~\ref{fig:assembly_demo_execution}).

\begin{table}[t]
    \centering
    \caption{\textbf{Printing dimensions and mating clearances.} Assembly identifies each AutoMate part pair in demonstration order, with plug and socket meshes shown in blue and gray, respectively. Plug overall size and Socket overall size give the respective mesh dimensions as $x\times y\times z$ in the original mesh frame. Hole $D_h$ is the effective bore diameter, measured by the largest cylinder that fits through the bore, including its narrowest ridges. Shaft $D_s$ is the effective shaft diameter, measured by the smallest cylinder enclosing the mating shaft segment. The diametral clearance is $\Delta=D_h-D_s$, and the conservative radial clearance $c=\Delta/2$ gives the available transverse offset for parallel mating axes. All dimensions and clearances are in millimeters. The hole is in the plug mesh for the first three pairs and in the socket mesh for \texttt{00446}.}
    \label{tab:assembly_print_dimensions}
    \footnotesize
    \setlength{\tabcolsep}{2pt}
    \renewcommand{\arraystretch}{1.2}
    \begin{tabular*}{\linewidth}{@{\extracolsep{\fill}}cccrrrr@{}}
        \toprule
        Assembly & Plug overall size & Socket overall size & Hole $D_h$ & Shaft $D_s$ & $\Delta$ & $c$ \\
        \midrule
        \raisebox{-0.5\height}{\shortstack{\includegraphics[width=19mm,height=13mm,keepaspectratio]{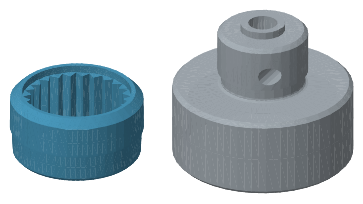}\\[-1pt]\texttt{00081}}} & $52.95\times52.98\times27.53$ & $71.59\times71.65\times58.26$ & 37.41 & 35.86 & 1.56 & 0.78 \\
        \addlinespace[4pt]
        \raisebox{-0.5\height}{\shortstack{\includegraphics[width=19mm,height=13mm,keepaspectratio]{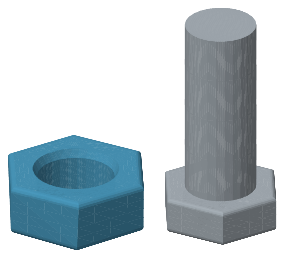}\\[-1pt]\texttt{00581}}} & $55.58\times48.32\times19.46$ & $45.52\times39.42\times78.83$ & 28.18 & 25.80 & 2.38 & 1.19 \\
        \addlinespace[4pt]
        \raisebox{-0.5\height}{\shortstack{\includegraphics[width=19mm,height=13mm,keepaspectratio]{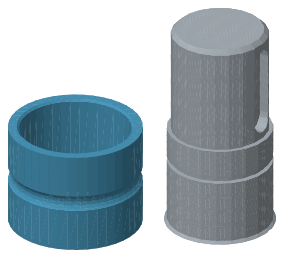}\\[-1pt]\texttt{00007}}} & $51.08\times50.94\times39.60$ & $42.42\times42.30\times81.44$ & 42.86 & 37.20 & 5.66 & 2.83 \\
        \addlinespace[4pt]
        \raisebox{-0.5\height}{\shortstack{\includegraphics[width=19mm,height=13mm,keepaspectratio]{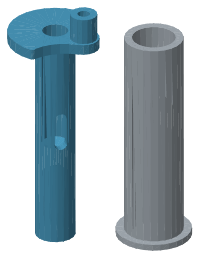}\\[-1pt]\texttt{00446}}} & $39.25\times39.16\times105.04$ & $31.32\times31.32\times90.81$ & 18.01 & 16.54 & 1.47 & 0.73 \\
        \bottomrule
    \end{tabular*}
\end{table}

\paragraph{Block construction from images.}
Six cubes are placed on a tabletop, and the robot arranges them into the structure shown in the goal photograph. The three configurations appear in Figure~\ref{fig:block_demo_execution}.

\paragraph{Action diversity.}
Dice flipping uses distinct initial and target faces balanced across predefined conditions. Grasp selection, rotation, and regrasping are annotated from video. For throwing, execution records and video establish joint motion, gripper opening, and object separation.

\paragraph{Bimanual towel folding.}
Each configuration provides an overhead demonstration video and images of the initial and final towel states. The agent may stage placements to maintain arm clearance. Final folds are assessed from overhead images after release and withdrawal. Figures~\ref{fig:towel_demo_execution} and~\ref{fig:towel_simultaneous_demo_execution} show the two configurations.

% \subsection{Metric Definitions and Resource Accounting}
% \label{sec:appendix_metrics}

% \paragraph{Success and time.}
% Success rate is $100s/n$, where $s$ is the number of completely successful trials among $n$ evaluated trials. The task criteria are specified in Table~\ref{tab:task_suite}. Completion time, token usage, and cost use the same successful trials, with sample counts and summary statistics reported in Tables~\ref{tab:task_comparison} and~\ref{tab:model_effort_assembly}. A configuration with no successes has undefined successful trial resource means.

% Completion time follows the task delivery to final physical verification boundary in Section~\ref{sec:exp_evaluation}. It includes program construction, observation, computation, motion, and recovery.

% \paragraph{Interaction accounting.}
% The index $k$ in Equation~\ref{eq:runtime-policy} counts agent tool decisions. Observation and action requests are counted individually, including those grouped within a single tool call.

% \paragraph{Token usage.}
% Total tokens sum input and output usage across all model requests in a trial, including retries. Repeated context counts on each request, with cached input counted once within that request's input total. Provider records are normalized so that reasoning contributes once to output. Table entries report mean total tokens per successful trial in millions.

\section{Supplementary Coding Agent Comparison on Pyramid Construction}
\label{sec:appendix_pyramid_agents}

Table~\ref{tab:pyramid_agent_comparison} extends the model comparison in Table~\ref{tab:model_effort_assembly} to another task (pyramid construction) using GPT-5.6 Sol and GPT-6 Astra at high thinking effort. This result echos the finding in Table~\ref{tab:model_effort_assembly} that GPT-5.6 is able to complete robotic tasks using AGP, while it also illustrates the capability differences between 5.6-Sol and Astra. 
% We use these two models and the pyramid construction task for illustrative purposes, and it does not imply that AGP is restricted to these two models or this task.

\par\addvspace{8pt}
\noindent\begin{minipage}{\linewidth}
    \captionsetup{type=table}
    \centering
    \caption{\textbf{Supplementary coding agent comparison on pyramid construction.}}
    \label{tab:pyramid_agent_comparison}
    \small
    \setlength{\tabcolsep}{4pt}
    \begin{tabular*}{\linewidth}{@{\extracolsep{\fill}}lrrr@{}}
        \toprule
        Model & Successes & Tokens (million) & Cost (USD) \\
        \midrule
        GPT-5.6 Sol & 7/10 & 18.86 & 9.22 \\
        GPT-6 Astra & 10/10 & 9.24 & 11.69 \\
        \bottomrule
    \end{tabular*}
\end{minipage}
\par\addvspace{8pt}

\paragraph{Usage and pricing.}
Inference costs are estimated using API prices retrieved on September 9, 2026.

\paragraph{Interpretation.}
Sol uses more tokens per successful session and has lower estimated inference cost than Astra.

\section{Runtime Program Analysis}
\label{sec:appendix_runtime_programs}

We analyze the programs constructed during 65 robot task executions to characterize the runtime programming described in Section~\ref{sec:method:programming}. These programs mainly serve three purposes: (1) perception and geometry, including depth measurement, color region selection, ray intersection, triangulation, surface fitting, and grasp offset estimation; (2) robot call composition, including pose and gripper commands followed by capture, concurrent calls to different arms, compact responses, and full response logging; (3) visual annotation, which projects spatial reference points into camera images for inspection. The logs also contain shell command events with inline Python, covering image processing and geometric computation as well as dependency checks, report writing, and file verification.

\paragraph{Case study 1: surface fitting and grasp offset compensation.}
In a four pair assembly execution, the agent creates \texttt{fit\_head.py} to estimate the pose of a held pin from a wrist depth image. It samples triples of points to identify a planar surface, then uses singular value decomposition to estimate its normal. The following excerpt refines the plane estimate from the selected surface points.
\begin{quote}
\small
\begin{verbatim}
ctr = best.mean(0)
_, _, v = np.linalg.svd(best - ctr)
normal = v[-1]
normal *= np.sign(normal[2])
\end{verbatim}
\end{quote}
The agent extends the program to intersect an overhead pixel ray with this plane, estimate the pin tip using an assumed pin length, and express the tip offset in the gripper frame. A further extension computes a corrected target orientation and position. The resulting horizontal target is approximately $(0.51325,-0.15820)$\,m. The next motion request uses $(0.5133,-0.1582)$\,m and the computed quaternion, establishing a direct link between the program output and physical action selection. That request returns a target arrival error. The agent then adjusts the requested position and continues the insertion sequence. This trace illustrates successive program extensions and the use of their outputs within a feedback loop.

\paragraph{Case study 2: triangulation from two camera views.}
In another four pair assembly execution, the agent creates \texttt{triangulate.py}. It converts selected wrist and overhead pixels into calibrated rays in the robot base frame. Given ray origins \texttt{t1} and \texttt{t2} and unit directions \texttt{d1} and \texttt{d2}, the program computes the closest points on the two rays using the following operations.
\begin{quote}
\small
\begin{verbatim}
ab = np.linalg.lstsq(
    np.stack([d1, -d2], 1),
    t2 - t1, rcond=None
)[0]
p1 = t1 + ab[0] * d1
p2 = t2 + ab[1] * d2
\end{verbatim}
\end{quote}
It returns their midpoint as the position estimate and their separation as a consistency measure. The agent subsequently calls the same program on additional targets.

\section{Task Examples}
\label{sec:appendix_task_illustrations}

\begin{figure}[p]
    \centering
    \includegraphics[width=\linewidth,height=0.84\textheight,keepaspectratio]{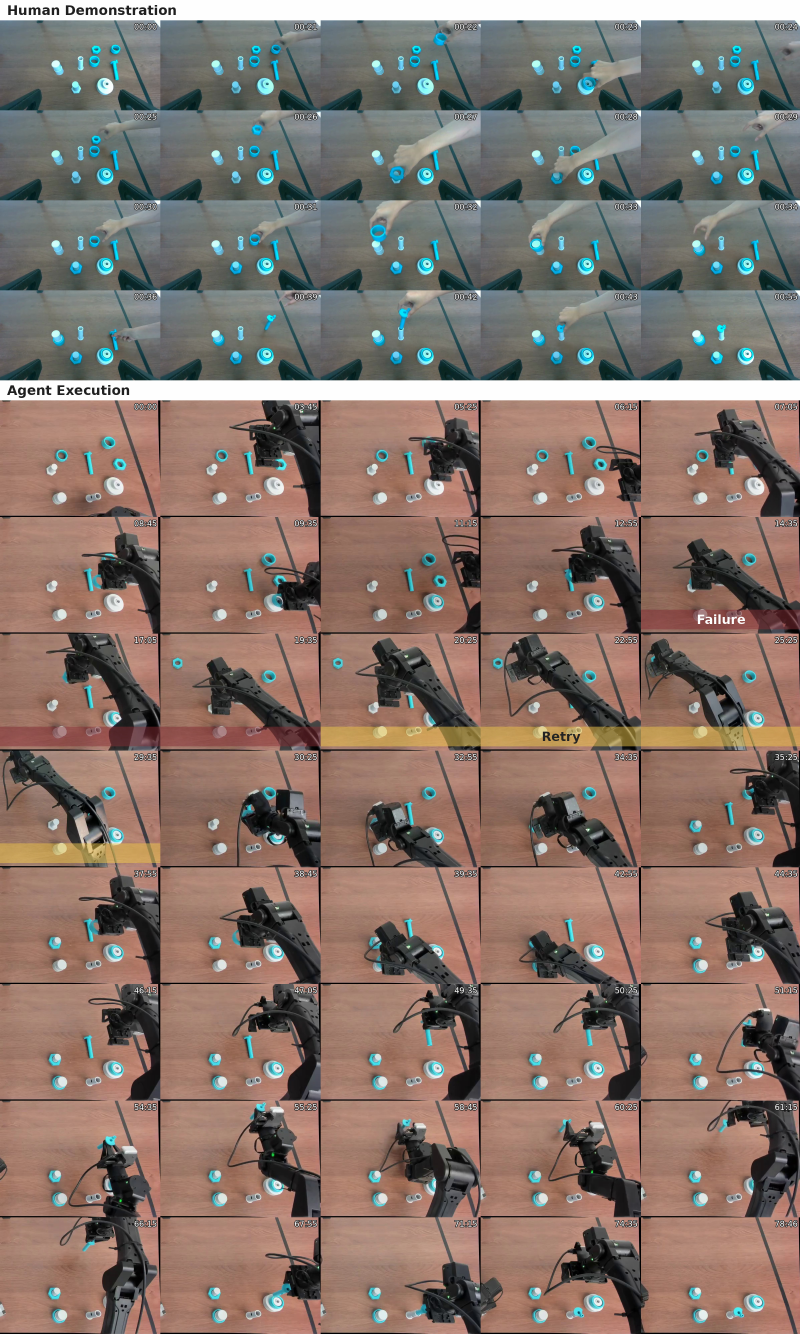}
    \caption{\textbf{Four pair assembly from a human demonstration video.} The upper panel shows four demonstrated assemblies. The lower panel shows the robot assembling eight parts into four pairs, illustrating video instruction following and long horizon reasoning. Dark red bars mark failed nut insertion and displacement. Yellow bars mark grasp retries.}
    \label{fig:assembly_demo_execution}
\end{figure}

\begin{figure}[p]
    \centering
    \includegraphics[width=\linewidth]{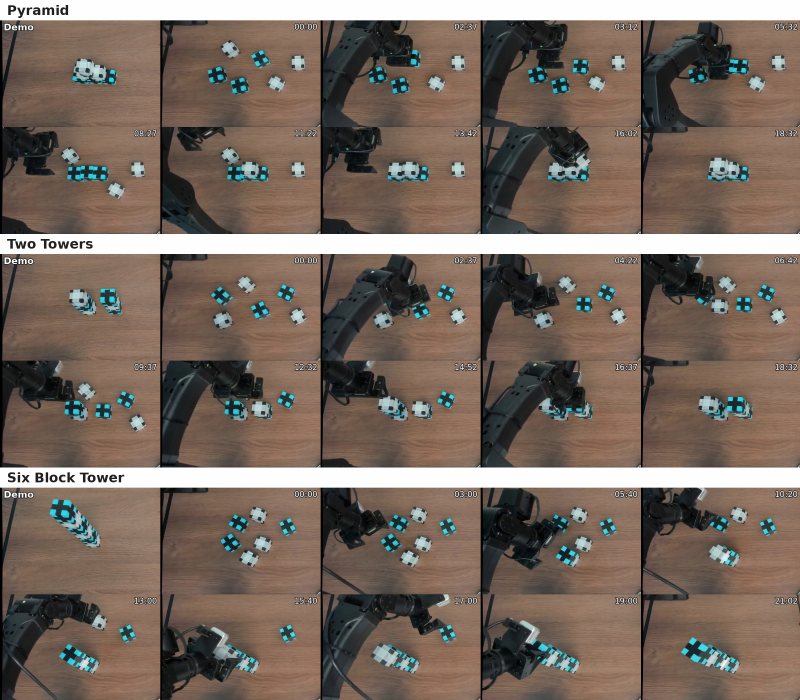}
    \caption{\textbf{Block construction from static goal images.} From top to bottom are the pyramid, two towers, and six block tower. Each occupies two rows of five images, beginning with the goal image labeled Demo and followed by nine execution frames.}
    \label{fig:block_demo_execution}
\end{figure}

\begin{figure}[p]
    \centering
    \includegraphics[width=\linewidth]{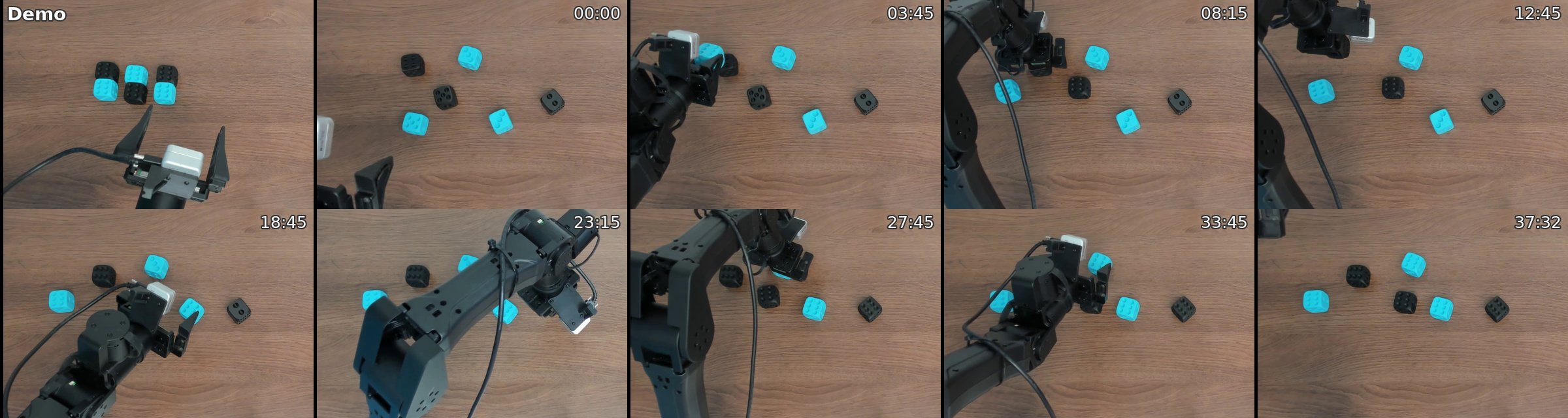}
    \caption{\textbf{Dice flipping across six dice.} The goal photograph in the cell labeled Demo specifies six pips upward on each of three black and three light blue dice. Nine overhead execution frames show the initial faces, successive manipulations, and final state with all six target faces upward. The task requires selecting grasps and rotations that produce the requested orientation after release.}
    \label{fig:die_flipping_demo_execution}
\end{figure}

\begin{figure}[p]
    \centering
    \includegraphics[width=\linewidth,height=0.84\textheight,keepaspectratio]{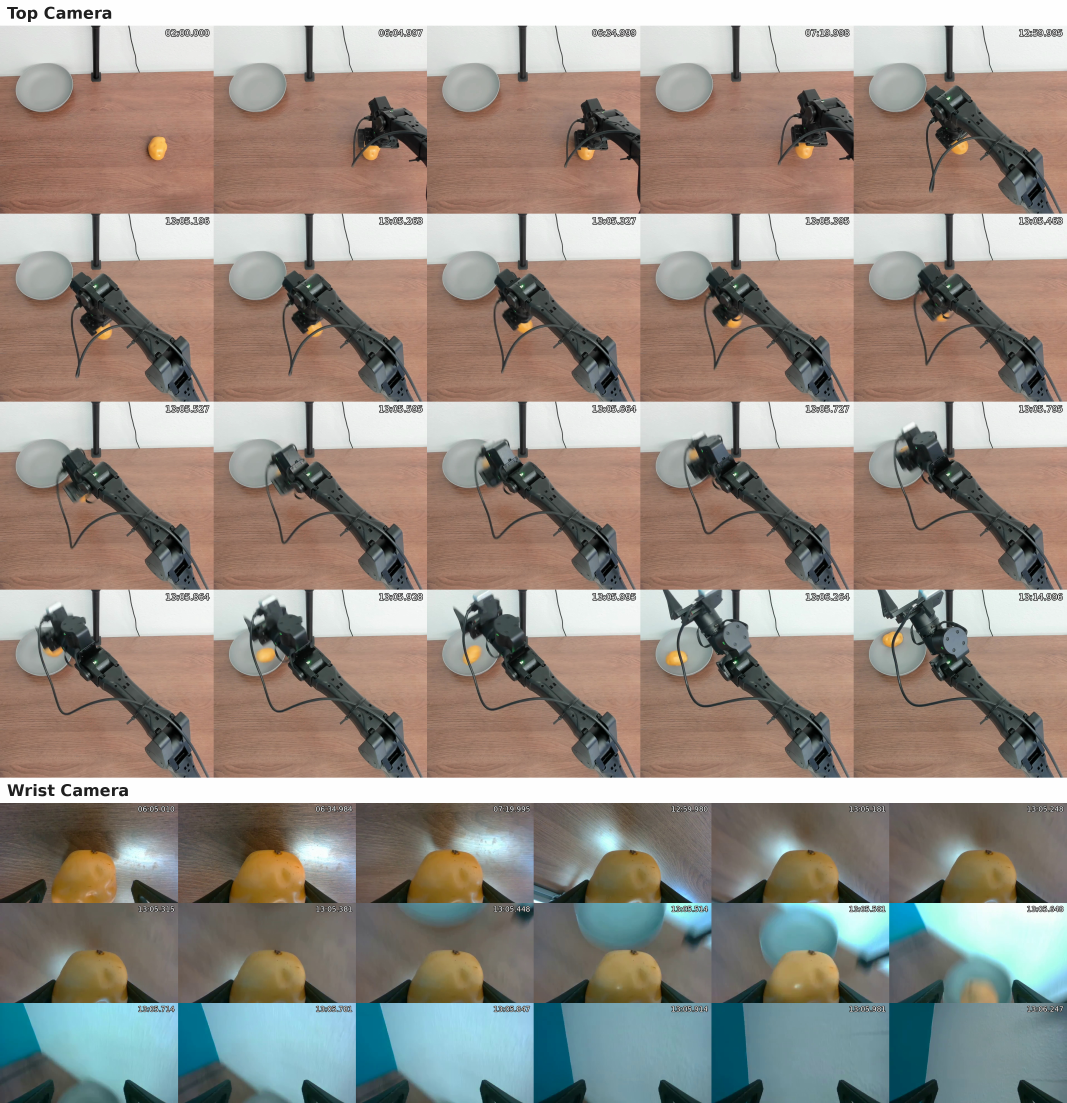}
    \caption{\textbf{Targeted throwing from a language instruction.} The task requires a circular swing and gripper release toward a bowl, illustrating the dynamic release component of action diversity. The overhead sequence shows the initial arrangement, grasp, swing, and final object position inside the bowl. The wrist sequence shows retention and separation at the gripper.}
    \label{fig:throw_execution}
\end{figure}

\begin{figure}[p]
    \centering
    \includegraphics[width=\linewidth,height=0.84\textheight,keepaspectratio]{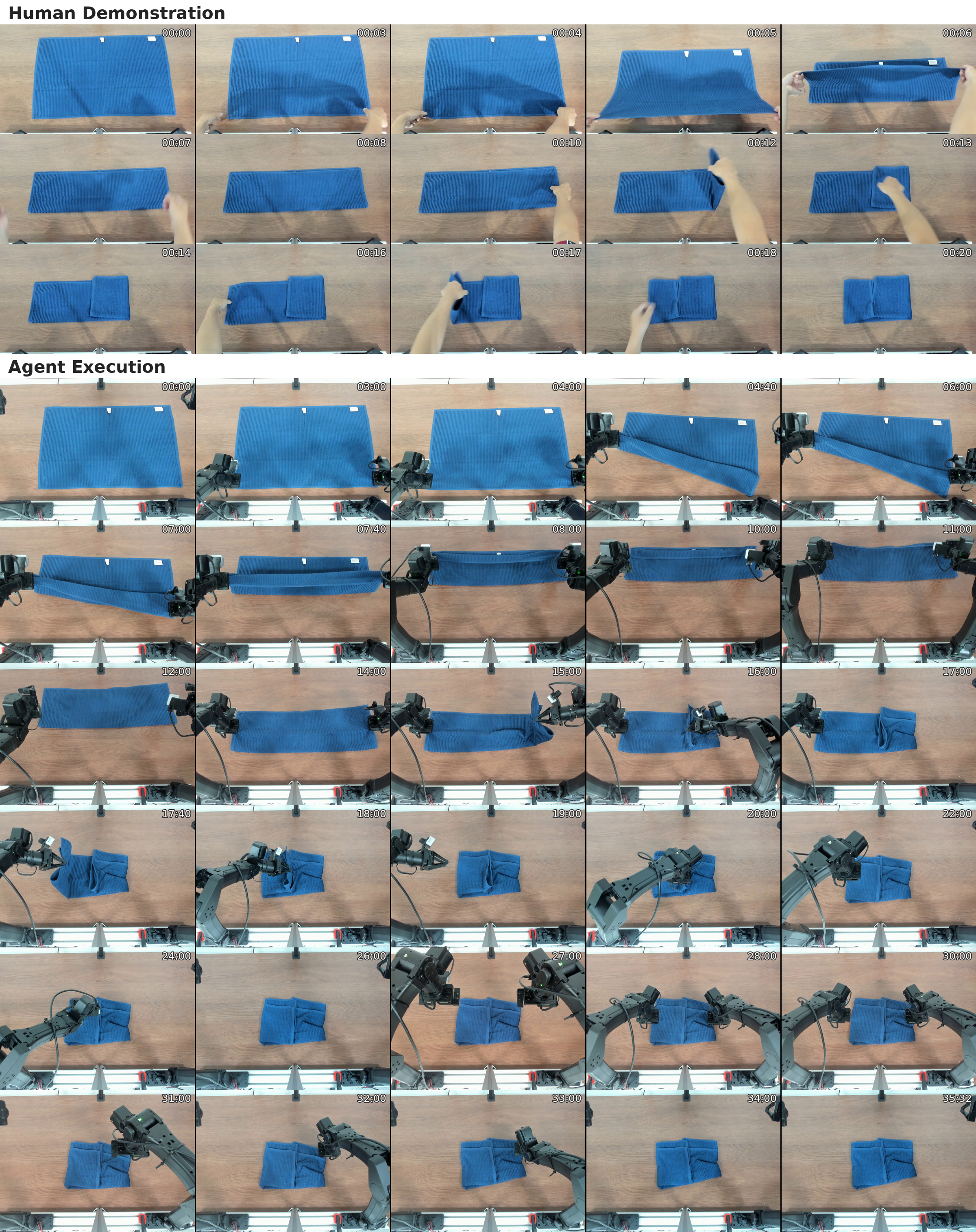}
    \caption{\textbf{Sequential bimanual towel folding from a human demonstration video.} The upper panel contains 15 frames showing the long edge fold followed by the right and left ends folding inward. The lower panel contains 30 frames showing coordinated grasping, recovery after a corner slips, the three folds, and subsequent adjustments.}
    \label{fig:towel_demo_execution}
\end{figure}

\begin{figure}[p]
    \centering
    \includegraphics[width=\linewidth,height=0.84\textheight,keepaspectratio]{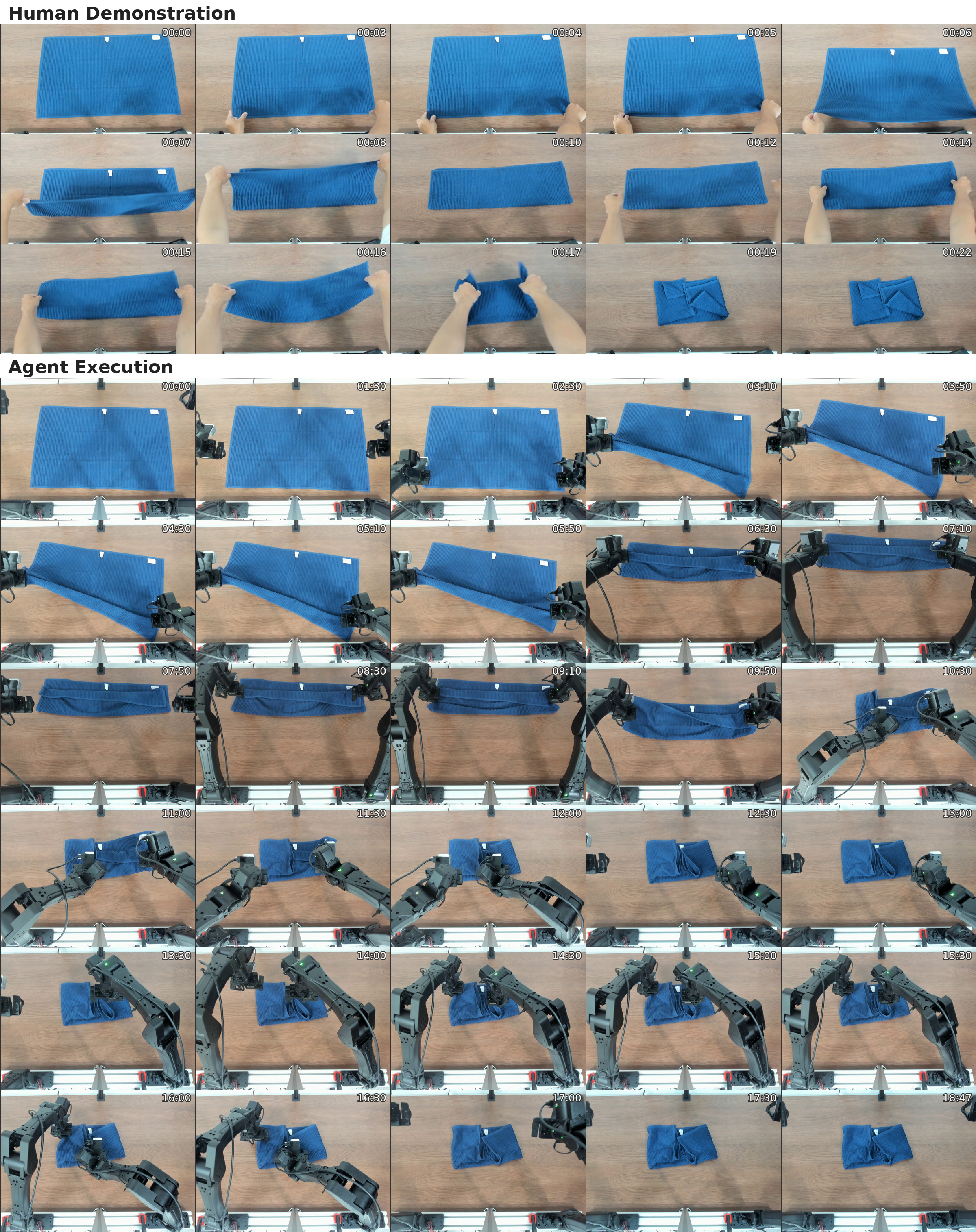}
    \caption{\textbf{Simultaneous bimanual towel folding from a human demonstration video.} The upper panel shows the long edge fold followed by both short ends folding inward together. The lower panel shows the robot execution, including recovery of a slipped corner, the long edge fold, inward transport of both ends, staged placement of the flaps, and alignment correction.}
    \label{fig:towel_simultaneous_demo_execution}
\end{figure}

Figures \ref{fig:assembly_demo_execution} to \ref{fig:towel_simultaneous_demo_execution} present video frames that illustrate AGP's completion of every task in Table~\ref{tab:task_comparison}.

\section{Open Weight Model Training}
\label{sec:appendix_open_models}

We use Slime to train Qwen3.6 on execution trajectories from GPT-6 Astra. Sequences contain at most 65,536 tokens. The batch size is 64, and the initial learning rate is $10^{-5}$ with 3\% warmup and cosine decay toward $10^{-6}$. The selected checkpoint is training step 175, at the end of the second epoch, which attains the lowest validation loss. Inference uses BF16 weights, with sampling at temperature 0.6, top-$p$ 0.95, and top-$k$ 20, and at most 32,768 output tokens per response.

\end{document}